# VALIDATION OF MASSIVELY-PARALLEL ADAPTIVE TESTING USING DYNAMIC CONTROL MATCHING

**Schaun Wheeler**

Aampe, Inc.

Raleigh, North Carolina, USA

schaun@aampe.com

**ABSTRACT**

A/B testing is a widely-used paradigm within marketing optimization because it promises identification of causal effects and because it is implemented "out of the box" in most messaging delivery software platforms. Modern businesses, however, often run many A/B/n tests at the same time and in parallel, and package many content variations into the same messages, not all of which are part of an explicit test. Whether as the result of many teams testing at the same time, or as part of a more sophisticated reinforcement learning (RL) approach that continuously adapts tests and test condition assignment based on previous results, dynamic parallel testing cannot be evaluated the same way traditional A/B tests are evaluated. This paper presents a method for disentangling the causal effects of the various tests under conditions of continuous test adaptation, using a matched-synthetic control group that adapts alongside the tests.

## 1 Introduction

Businesses are rarely able to invest in all of the opportunities available to them. A business may not have the resources to invest in all options; investing in one option may preclude investment in an alternative;

and those responsible for exercising the options may simply not have enough time or attention to do everything. Because of these constraints, businesses often look to select the subset of all available options that will maximize their return on investment. The ability to do optimized subset selection is arguably a major determinant of how well the business does, and how quickly it can grow.

Marketing communication is subject to constraints that make optimized subset selection necessary. (A business that communicates with all customers at all times about all things will lose all of its customers, and incur unmanageable message delivery costs in the process.) However, selecting the optimal messaging parameters – timing, topic, copy details, and so on – strains marketing decision-makers' ability to make good decisions. It is labor-intensive to identify available options, set up evaluations of each, parse results, and transform those results into subsequent operational decisions that themselves should be subject to evaluation.

Most marketing communication delivery platforms for email, SMS, push notifications and other channels include the ability to conduct A/B tests. The premise and process of A/B testing is simple: (1) choose two options to act as treatments in the experiment, (2) randomly allocate a sample of contacts (customers, app users, etc.) to each treatment, (3) run the treatments - in other words, send the messages to the contacts, (4) measure open rates, click rates, add-to-cart rates, purchase-rates, retention rates or any other metric(s) that matter to the business. The treatment with the better success metrics is the better option.

There are many different "best practices" for determining by how much the winning option's results must exceed the losing option's results for the experimenter to conclude that the winning option really is more effective. These validation practices, however, are often premature, as A/B testing, in its most commonly-implemented form, fails to account for a wide variety of sampling biases and instrumentation biases. The most prominent of these are (1) asymmetric allocation of high-value contacts to one specific treatment just by random chance [1, 2], and (2) failure to account for variations in the treatment necessarily introduced by delivering the treatment in the form of a human-readable message [3]. Additionally, even robust test results can prove to have short-term value because of seasonality, user adaptation, or other effects [4, 5]. In short, A/B testing makes it very easy for a business to distort or misinterpret test results.

A variety of approaches derived from Reinforcement Learning (RL) - most notably multi-armed bandits and contextual bandits - can address the inherent weaknesses in A/B tests while still retaining their useful inferential properties. An RL approach can monitor all messaging choices that might influence contact behavior, and monitor the subsequent behaviors themselves to automatically select high-impact choices for subsets of contact population. Because an RL approach is inherently adaptive, and because it inherently considers many different possible influences at the same time, it mitigates the problems of asymmetric allocation of treatments, confounding and/or moderating treatments, and test result durability.

One challenge of any continuously-adaptive system is validation, as such systems have no "before" and "after" period in which to compare results. To address this difficulty, we present a method for compiling an adaptive control group. We first provide a short overview of one way a RL system can be implemented in practice to adaptively optimize messaging in an app user ecosystem, thereby generating systemic impacts. We then evaluate some of the other methods traditionally used for measuring systemic impacts, and explain how an adaptive control overcomes shortcomings in those methods. We then present the guidance for constructing the control, and show results from one corporate user of a real-world RL messaging system.

## 2. One way to implement a reinforcement-learning messaging pipeline

The following is a short summary of how we have implemented a continuously-adaptive messaging pipeline based on reinforcement learning rather than A/B tests.[1] We have provided a full justification of that learning architecture elsewhere [6]. This is only a high-level overview, and of course, it is only one possible way to construct such a system.

### 2.1 Representation of general behavioral patterns.

The pipeline must first have the ability to monitor user behavior. This means the system can sit on top of an app or website's events stream, using only first-party data. For an e-commerce app or site, instrumented events may include activities like logging in, updating a profile, viewing a product, adding a product to cart, starting the checkout process, completing a purchase, and so on, but could also include low-level events such as the click of a specific button or the viewing of a particular recommendation.

One of the most important features of first-party data is that these events all come not only with a timestamp but also a persistent id designating a unique contact. This means event summaries can be calculated for every contact for different time horizons (one week, one month, etc.) to summarize what "normal" behavior for each contact looks like. If necessary and/or desirable, dimensionality reduction techniques such as Principal Components Analysis can provide a condensed representation of each contact's activity that lessens the computational complexity of later analysis.

### 2.2 Intermeshed treatment assignment

---

[1] https://aampe.com

An RL-based learning system does not need to identify discrete "tests" of only a handful of treatments.[2] In real-world messaging situations, such tests are not feasible anyway, as treatments must be delivered to contacts in human-readable packages of words. For example, a message may include a value proposition that emphasizes convenience, and employ a minimalist, business-like tone; another message may include the same value proposition, but use a casual, encouraging tone; another message may include a value proposition that emphasizes low cost, but uses the business-like tone; and so on. Writing messages to accommodate all of the possible combinations of treatments can be difficult to do manually. However, it is possible to create multiple text snippets that conform to each desired treatment, and then combine snippets to ensure that different combinations of treatments are available to be sent to contacts.[3]

When the system has a high-confidence estimate in a treatment's expected performance, it delivers that message more frequently.[7] To the extent that clearly better-performing estimates do not exist, the system selects a message that explores the treatment space more fully. In case of both exploration of new possibilities and exploitation of past learnings, random perturbations to the decision process prevent optimization to local maxima.

## 2.3 Behavior monitoring and rewards definition

Much like a standard A/B test, an RL system can use a predetermined monitoring window (say, the 24 hours following the send of a message) to identify which contacts performed which actions on the app. If desired, different events can yield different amounts of reward to the system, and the time between message and event can be used to discount rewards. For example, assuming that the most desired event for the system is for a contact to complete a purchase, completing a purchase within one hour might receive a high reward weight; visiting the app within an hour but not completing a purchase might receive a lower weight; a a purchase eight hours after the message might also receive a lower weight; visiting the app after eight hours and not completing the purchase might receive the lowest weight of all.

## 2.4 Model-based inference

The system we reference here optimizes at the individual contact level - essentially a contextual bandit where each user is a separate context. Therefore, inference takes place at the level of unique user ids. In a more traditional contextual bandit, this inference could take place for each context, even if that context was a relatively large segment of the total population. However, for the sake of simplicity and clarity in the following discussion, we will assume that the context is the individual user.

---

[2] In our experience, companies who implement multi-armed bandits often retain the practice of handling different messages as discrete treatments, perhaps because it is easier to think about and evaluate performance when doing so. We believe this leaves a lot of value unrealized, as intermeshed treatments allow for inference that enables cross-selling and other applications of results from one message to other messages.
[3] See, for example:
https://www.aampe.com/blog/how-to-get-more-engagement-from-your-existing-customers-without-making-any-new-content

One simple way to model the type of dynamic system described above is to create a training set incorporating historical behavior (or a latent-space representation thereof), the treatments received (even if multiple treatments were packaged in the same message), and previous model estimates of each contact's treatment preferences. Training data can be weighted by the reward weights, if such weights were calculated. This can then be modeled as a classification problem: positive classes are cases where messages resulted in a goal event being achieved, and negative labels are assigned to messages where goal events were not achieved. If reward weights were used, these weights can be thresholded to create positive and negative classes.

The benefit of using a model is not only that it can incorporate a wider variety of information than a human operator could reasonably digest, but is also allows for the estimation of two different predictions: (1) a baseline probability of goal achievement based only on the historical features, and (2) a marginal probability for each possible treatment, where that treatment and the contact's preference for the treatment are included in the model alongside historical behaviors. The target estimates are then compared to the baseline estimates to estimate the probability that a contact will achieve a goal, *above baseline expectation*, if exposed to a particular treatment.[4]

## 2.5 Feed-forward assignment

These estimates of user preference can be stored in the system and used for subsequent treatment assignment and evaluation. Learned preferences can be used to make choices for timing (day of week, time of day, etc.), topic (product category, brand, etc.) copy (value proposition, call to action, tone, etc.) and channel (push notification, email, SMS, in-app notification, etc.), as well as any other ways an operator might want to vary their messaging.

## 3. Holdout groups and systemic impacts

The system outlined in the previous section is a continuously-adaptive alternative to time-bound A/B tests. As such, evaluation of such a system is more challenging than evaluating a simple choice between A and B. It is also more challenging than evaluating the impact of a one-time intervention such as a policy change. In this section, we review three methods for estimating systemic impacts, and discuss why none of the methods are appropriate for evaluating a dynamic messaging system.

**Global holdout.** The simplest form of control group is a holdout sample. In this method, a subset of contacts would be held out from messaging entirely. Creating a holdout that remains comparable to the other records is difficult [8]. Also, the larger the holdout - and it must be sufficiently large to be analytically useful - reduces the benefits that can accrue from experimental intervention [9]. Global holdouts are not appropriate for assessing dynamic systems. In the context of marketing messaging, a global holdout, if properly curated and maintained, would allow conclusions about the benefit of messaging relative to never being messaged ever. That is not a realistic business case. The important question is whether messaging can improve upon whatever the status quo is at the moment. In adaptive, iterative analysis, a global holdout can lead to overfitting on the holdout [10].

---

[4] There are many different ways to justifiably construct a model to estimate the propensity-above-baseline to engage in a particular reward behavior. Evaluating those different modeling choices is outside the scope of this document. For more information on one possible set of choices, see https://www.aampe.com/blog/how-do-i-know-it-works-technical-explanation

**Switchback testing.** Switchback testing alternates the time period in which contacts receive treatments [11, 12, 13]. So, for example, half the contacts may receive one message on one day, and the other half receive nothing. The next day, the previously-unmessaged half gets the message the first half received, and the first half gets nothing. While switchback testing is an effective way of controlling for network effects, it cannot detect learning effects. In an adaptive system, learning effects are one of the main effects we expect [14].

**Synthetic control.** A synthetic control identifies predictors of an outcome that will not themselves be influenced by the treatment. A model is fit to predict the outcome based on those factors [15]. The treatment is then administered, and the difference between the actual outcome and the predicted outcome is the estimate of the causal effect. Synthetic controls are a powerful tool, but they exhibit two marked drawbacks. The first is that the estimate of causal effects is dependent upon the identification of independent predictors and the accurate modeling of the outcome based on those predictors. Such predictors may not always be available, and goodness of model fit is always a subject of debate [16]. The other problem is that synthetic controls, similar to a global holdout, are able to model change from some initial, unchanged state, but cannot easily model a change from the most recent state in a constantly-changing system. They are more appropriate for one-time interventions.

Available methods for constructing control groups to estimate causal impact of interventions all show definite strengths, but they fall short in addressing the kind of dynamic, adaptive system described in this paper.

## 4. Adaptive control via coarsened exact matching

Our approach to control groups takes three principles from standard holdout practices: (1) the control group must not have been exposed to the treatment, (2) time windows can be used to bound exposure, and (3) control cases can be explicitly selected based on their similarity with treated cases to ensure comparability.

Coarsened Exact Matching (CEM) is a statistical method used in causal inference studies to reduce sampling bias created by non-random sampling [17]. Working on the theoretical basis as stratified random sampling [18], CEM creates approximate matches between groups of similar individuals from different study populations. Individuals who received a treatment are matched to users who did not receive a treatment based on these strata.

We employ CEM to create a synthetic control group that adapts as the test population adapts. Theoretically, any combination of factors could be used to do the matching. We have successfully implemented a control that takes the following factors into account:

1. **Time window.** Each message sent is matched to a user who (1) received a message 2-8 days before the message in question, and (2) did not receive any message between 1 day before and 1 day after the message in question. The first criterion reduces seasonal effects by matching on a reasonably recent message, while the second criterion reduces instrumentation bias that would result from multiple treatments being administered and evaluated simultaneously.
2. **Message clusters.** Messages can be clustered based on word similarity, on which treatments were embedded in the message, or any other characteristics. The point is to capture some measure of similarity in messages so contacts who have been assigned a particular message will be matched with contacts who recently received the same type of message.
3. **User activity level.** It is possible to categorize app or site users based on any number of features. In practice, we have found it useful to categorize users based on (1) how recently they experienced a "conversion" event - in e-commerce, this would be a purchase, and (2) how recently they visited the store or app in general. Each of these two activities is categorized into "active" behavior (occurrence in last 7 days), recent behavior (last 8-30 days), and "lurking" behavior (last 31-90 days). There can also be a category for new users (first seen on app or site in the last 7 days).[5] Matching on these categories ensures that users will only be matched to other users who have similar baseline levels of behavior.

Once a control user has been matched to a message, app and site behavior of both contacts can be monitored (though only the contact of the test message would actually receive any communication).

---

[5] The day thresholds listed here are notional. Any threshold could be used.

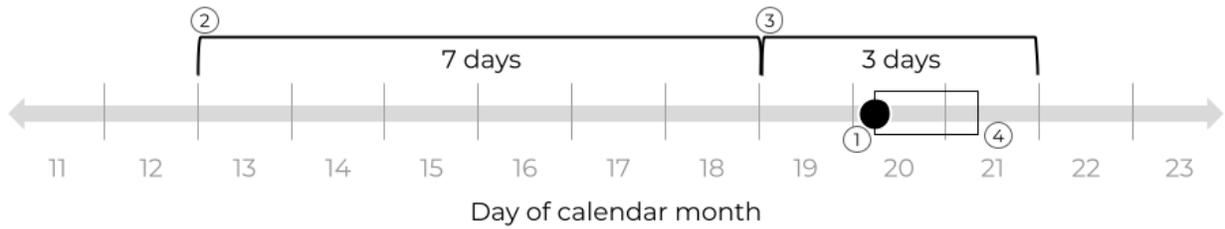

Figure 1: Time Window Example. As a default, all control candidates are selected in reference to a test message (1). Control candidates are selected if they received a message within the 7-day period preceding the day before the test message was sent (2), and did not receive a message the day before, during or after the test message was sent (3). Effects of both the test message and the matched control are monitored for 24 hours after the test message is sent (4).

The strength of the CEM approach, aside from the statistical properties of a matched test and control, is that it does not require a model, and therefore findings do not need to be tempered by goodness-of-fit measures. CEM results are unreliable when strata do not contain enough samples to match [19], but strata contents can be easily checked, and strata revised to ensure matches exist.

## 5. Results case study

The following results are from a large e-commerce app. We employed our default definitions for user activity categories:

1. New: seven days or less on the app or site.
2. Active: seven days or less since last conversion or visit event
3. Recent: Eight to 30 days since last conversion or visit event
4. Former: 31 to 90 days since last conversion or visit event

Because CEM does not rely on a model, goodness-of-fit metrics are not a consideration. However, it is reasonable to want some assurance that the matches are appropriate. Table 1 shows two different ways of evaluating match quality.

Table 1: Match quality metrics

| Category | Subcategory | Messages | | | Contacts | | |
|---|---|---|---|---|---|---|---|
| | | Test | Control | Coverage | Test | Control | Coverage |
| New | New | 259,076 | 234,633 | 90.6% | 36,125 | 2,208 | 6.1% |
| Active | Actively converting | 242,983 | 242,949 | 100.0% | 61,782 | 28,856 | 46.7% |
| Recent | Recently converted | 393,930 | 393,924 | 100.0% | 98,212 | 70,677 | 72.0% |
| Lurking | Formerly converted, actively looking | 234,056 | 234,055 | 100.0% | 85,194 | 52,682 | 61.8% |
| Lurking | Formerly converted, recently looked | 54,642 | 54,642 | 100.0% | 35,844 | 28,282 | 78.9% |
| Lurking | Actively looking | 1,698,175 | 1,698,175 | 100.0% | 443,722 | 296,271 | 66.8% |
| Lurking | Recently looked | 791,884 | 791,884 | 100.0% | 375,422 | 312,630 | 83.3% |
| Inactive | Formerly converted, formerly looked | 10,857 | 10,852 | 100.0% | 7,194 | 6,202 | 86.2% |
| Inactive | Formerly looked | 193,846 | 193,846 | 100.0% | 109,178 | 92,568 | 84.8% |

"Test" messages are messages actually sent to contacts. For each of those messages, we search for a list of viable control users, employing the methods outlined in the previous section. We randomly select one of those control users and the system logs a "Control" message for that user, recorded as being sent at the same time as its matched test message, but in actuality not being sent at all. The percentage of test messages for which we locate at least one control is one measure of the quality of the CEM procedure. In most cases, this percentage is near 100%. Notice, for new users, it is a little lower, because the window for new users is very small.

Another way to look at the quality of the CEM procedure is to take the radio of control contacts to test contacts. Over an arbitrary period of time (we used 14 days in the table above), a single contact may be messaged many times - under the settings for the system we evaluate here, as many as 35 times in a single week, but usually closer to 2 or 3 times in a week. For this particular case, the ratio of control to test contacts for new contacts is quite low. This means that the same control contact is being randomly matched to many different test contacts, which suggests the pool of control contacts for new users is quite low. A look at the raw numbers show that there are still over 2,000 test contacts for new contacts. That is a reasonably large sample size for computing summary statistics. Still, test/control comparisons for new users for this particular case should probably be treated with more caution than comparisons for other groups.

Because the system runs continuously evolving experiments rather than discrete, time-bound A/B tests, there is no single point at which we can simply gather "results" to show performance of test records relative to control records. In the system presently under consideration, messaging decisions are made based on decision confidence scores that summarize individual contact preferences for different message timing or copy decisions. In other words, every messaging decision is a test of the hypothesis that the system has accurately estimated each contact's messaging preferences.

We can use these scores to gauge relative performance dynamically by binning messages into reasonably sized samples based on the messages' confidence scores. So the lowest scores go into one bin, then the next lowest scores go into the next bin, and so forth until the highest scores go into the highest bin. The bin size is arbitrary - we generally use a bin size of 1000 messages.[6] Once messages are binned, we can calculate any performance metric we wish, and see an estimate of test vs. control[7], as shown in Figure 2.

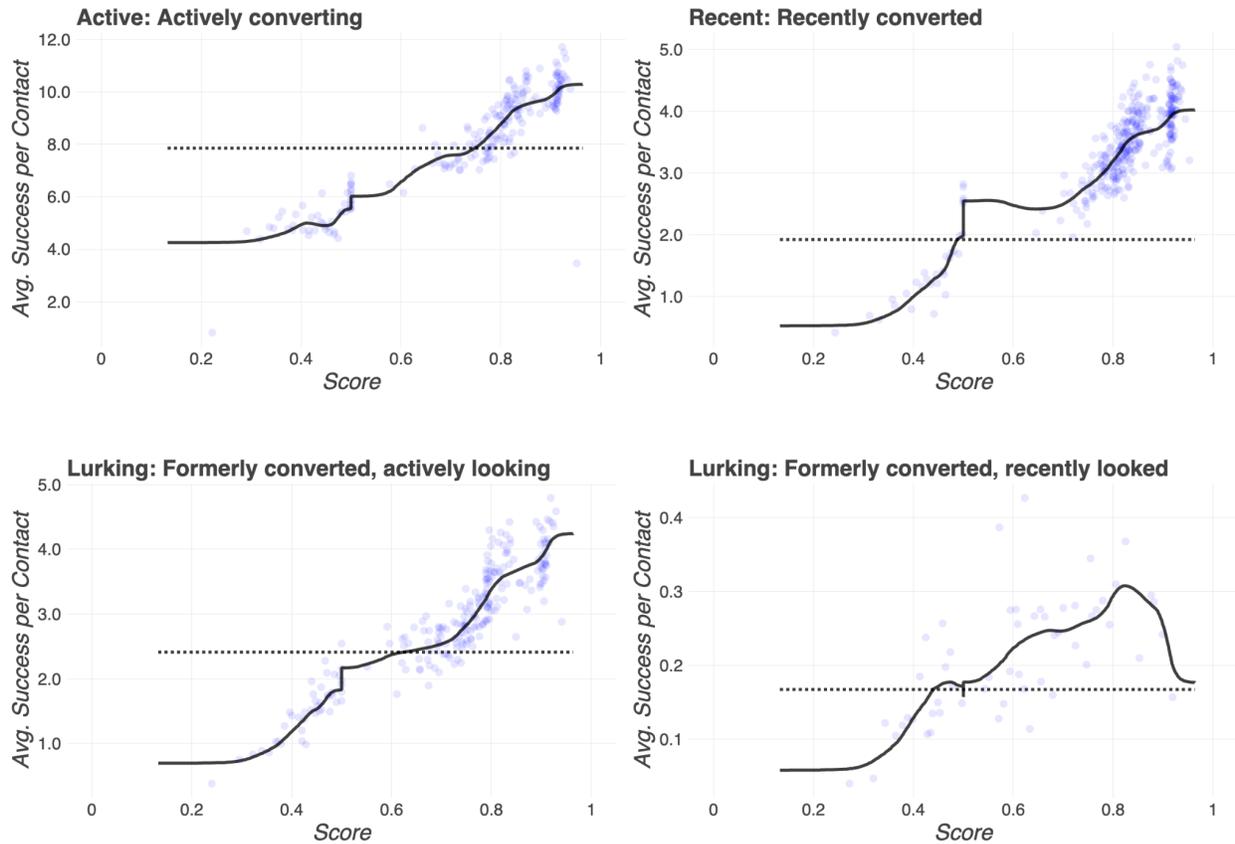

Figure 2: Average number of app/site visitation events per contact per bin for four different activity categories. Dotted line shows control values. Blue dots test values. Solid line shows smoothed test values.

As shown, the control metrics (the dotted lines) remain constant across bins. This is what we would expect from a control set, as control contacts were matched randomly to test messages. Generally, the higher the confidence score, the greater the extent to which the test messages outperform the control messages. This is what we would expect if the confidence score heuristics did in fact capture information about messaging decisions that make contacts more likely to respond.

---

[6] The number of messages to be binned cannot be assured to be a multiple of 1000. If a bin has less than 1000, then we treat the difference between 1000 and the actual count as a prior when calculating success metrics. The further the actual count is from 1000, the more we shrink the metric to 0. So if a success metric was 100% based on a bin continuing only 100 samples, then the revised metric would be (100% * 0.1) + (0% * 0.9) = 10%.

[7] We generally used as a success metric the average number of app or site events a contact triggered within 24 hours of being messaged. The default events we track are app visits (any visit to the app or site), adds to cart (a strong signal of intent that is, nevertheless, not a conversion event), and revenue generation (an actual conversion event). These names are adapted to e-commerce use cases, but non-e-commerce use cases usually have analogous events they want to monitor.

For three of the four categories shown in Figure 2, test messages start to outperform control messages at about the 0.5 mark. This is also expected, given the 0.5 is the starting value for confidence scores, indicating that a contact has neither a positive nor negative preference regarding a particular messaging decision, giving them a 50/50 chance of responding to such a decision. However, in the case of "active" users (those that did a conversion event within the past 7 days prior to the message), the test messages do not start to outperform the control until a much higher score bin - nearer to 0.75. That is because active users are already in a state of doing the desired event. Their baseline is higher, therefore it takes a message with stronger influence to move a contact above that already-high baseline.

Figure 3 shows the same results, but instead of evaluating messaging impact on general app or site visitation, we show the impact on revenue-generating events (checkouts completed).

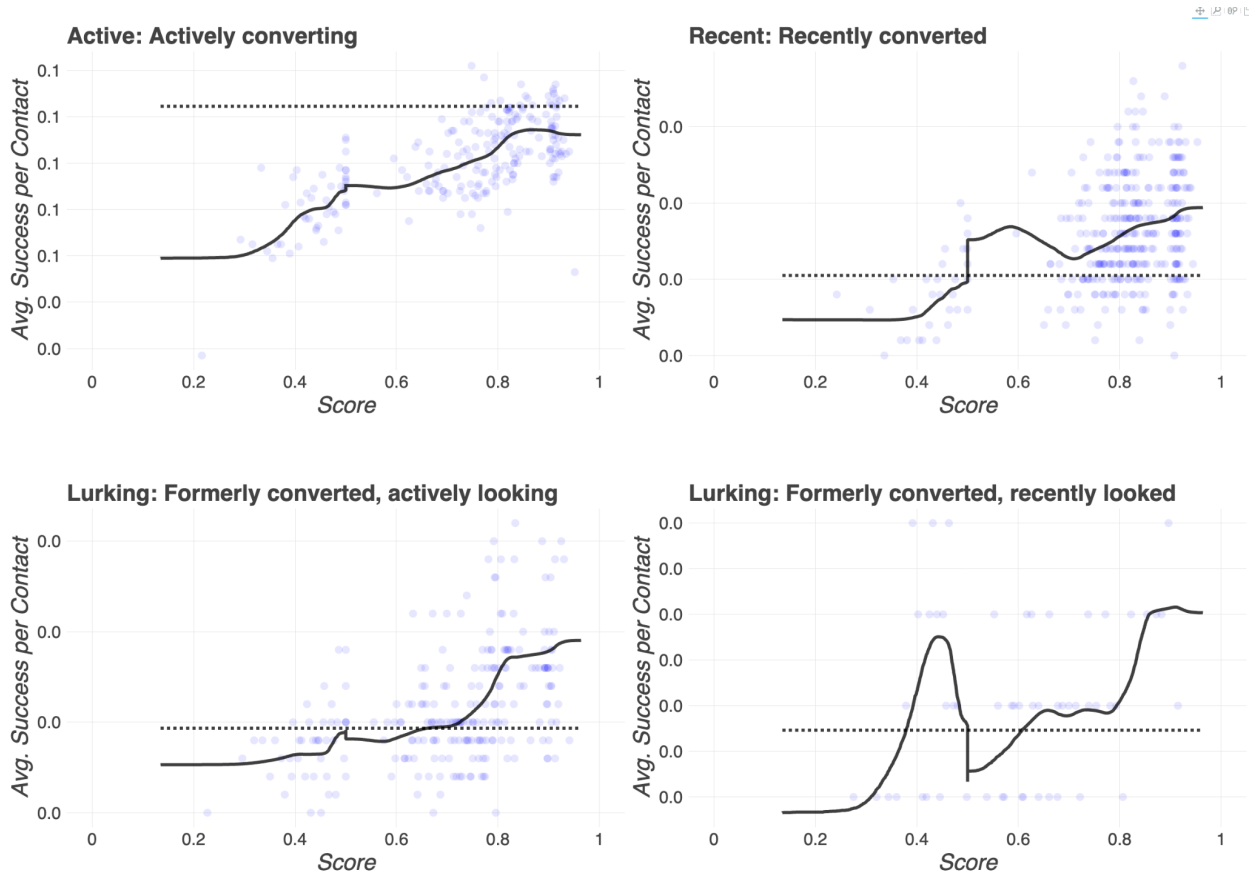

Figure 3. Average number of checkout-completed events per contact per bin for four different activity categories. Dotted line shows control values. Blue dots test values. Solid line shows smoothed test values.

These results show that, perhaps unsurprisingly, success metrics can look very different and perform differently depending on how you define success. Dynamic control groups can be used to evaluate messaging impact on any kind of contact behavior so long as that behavior does not itself depend upon a message being sent. For example, the dynamic control group cannot measure impact on click-through-rates, because contacts who are sent no messages by definition cannot click on a message.

## 6. A note on attribution

Nearly all businesses care about attribution of marketing or advertising efforts. Dynamic control groups allow for straightforward estimation of attribution. Each score bin represented in Figures 2 and 3 contains 1000 messages, but also contains a number of successes (app or site visits, adds to cart, purchases completed, etc.) observed within 24 hours of sending those messages. If the control success metric is higher than the test metric in a bin (in the figures: if the dotted line is higher than one of the blue dots), then obviously no credit can be assigned to the RL algorithm for that success. However, if a success metric is higher than the control metric, then the amount by which the test exceeds the control can be attributed to messaging.

For example, if a bin shows an average of 4.3 app visits for test messages, and 1.7 app visits for control messages, then that means that (4.3 - 1.7) / 4.3 = 60% of the success seen from test messages in that bin can be attributed to the messages rather than the random chance or baseline expectations. Table 2 shows attributable percent of successes for each user activity category.

Table 2: Percent of Success Attributable to Messaging

| Category | Subcategory | App visited | Item added to cart | Revenue Generated |
|---|---|---|---|---|
| New | New | 5.6% | 23.5% | 1.0% |
| Active | Actively converting | 10.6% | 6.5% | 0.3% |
| Recent | Recently converted | 40.9% | 37.6% | 29.4% |
| Lurking | Formerly converted, actively looking | 18.6% | 17.5% | 19.8% |
| Lurking | Formerly converted, recently looked | 21.7% | 25.6% | 31.2% |
| Lurking | Actively looking | 11.8% | 12.1% | 20.0% |
| Lurking | Recently looked | 11.8% | 15.1% | 15.9% |
| Inactive | Formerly converted, formerly looked | 0.0% | 7.9% | 12.0% |
| Inactive | Formerly looked | 6.0% | 9.7% | 7.6% |

As the table shows, messaging, for this use case, had the greatest impact on users who had recently converted but were not actively converting at the time of the messaging. Messaging also had a substantial impact on users who converted relatively long ago, but had relatively recently visited the site or app. The impact also varied across success metrics. Messaging did much more to influence already active users to visit the app than it did to influence purchases, but for those users who had purchased long ago and had recently visited the app, messaging had more impact on purchase behavior than visit behavior.

Multiplying the attributable percentage numbers by the actual count of successes in each bin, and then summing across bins, we can estimate the incremental number of visits, adds-to-cart, or purchase events attributable to messaging. All of this depends upon the selection of a monitoring window within which events will be considered candidates for attribution.

## 7. Conclusion

A/B testing is, in practically every way, inferior to dynamic intervention decisions based on reinforcement learning. However, A/B testing is not only ubiquitous, it also is very easy to generate easy-to-interpret results. For reasons discussed at the beginning of this paper, these results are often spurious or inflated, but it is easy for non-technical operators to misunderstand, downplay, or deliberately ignore that fact in light of the need to justify day-to-day decisions to their co-workers, managers, and employers. Historically, dynamically-adjusted experiments have been difficult to evaluate except by using a global holdout and monitoring for long-term effects (essentially turning the entire dynamic system into one huge A/B test). The dynamic control assignment based on coarsened exact matching demonstrated in this paper shows that these systems can be monitored for immediate effects, and generate meaningful attribution estimates as well.

## References

[1] Wang, Yu and Gupta, Somit and Lu, Jiannan and Mahmoudzadeh, Ali and Liu, Sophia. On Heavy-user Bias in A/B Testing. *arXiv preprint arXiv: 10.48550/ARXIV.1902.02021, 2019*.

[2] Davidson-Pilon, Cameron. The Class Imbalance Problem in A/B Testing. *https://dataorigami.net/2015/11/22/The-Class-Imbalance-Problem-in-A-B-Testing.html*

[3] Wang, Jingshu and Zhao, Qingyuan and Hastie, Trevor and Owen, Art B. Confounder Adjustment in Multiple Hypothesis Testing. *arXiv preprint arXiv: 10.48550/ARXIV.1508.04178, 2015*.

[4] Deng, Alex and Li, Yicheng and Lu, Jiannan and Ramamurthy, Vivek. On Post-selection Inference in A/B Testing. In Proceedings of the 27th ACM SIGKDD Conference on Knowledge Discovery, arxiv: 10.1145/3447548.3467129, 2021.